\documentclass[10pt,twocolumn,letterpaper]{article}

\usepackage{cvpr}              % To produce the CAMERA-READY version
\definecolor{cvprblue}{rgb}{0.21,0.49,0.74}
\usepackage[pagebackref,breaklinks,colorlinks,allcolors=cvprblue]{hyperref}
\usepackage[accsupp]{axessibility}
\usepackage[table]{xcolor}

\def\paperID{3187} % *** Enter the Paper ID here
\def\confName{CVPR}
\def\confYear{2026}

\title{Evaluating Generative Models via One-Dimensional Code Distributions}

\author{
Zexi Jia\textsuperscript{1},
Pengcheng Luo\textsuperscript{2},
Yijia Zhong\textsuperscript{3},
Jinchao Zhang\textsuperscript{1}\thanks{Corresponding author.},
Jie Zhou\textsuperscript{1}
\\
\textsuperscript{1}WeChat AI, Tencent Inc., China\\
\textsuperscript{2}School of Intelligence Science and Technology, Peking University\\
\textsuperscript{3}College of Computer Science and Artificial Intelligence, Fudan University
}

\begin{document}
\maketitle

\begin{abstract}
Most evaluations of generative models rely on feature-distribution metrics such as FID, which operate on continuous recognition features that are explicitly trained to be invariant to appearance variations, and thus discard cues critical for perceptual quality. We instead evaluate models in the space of \emph{discrete} visual tokens, where modern 1D image tokenizers compactly encode both semantic and perceptual information and quality manifests as predictable token statistics. We introduce \emph{Codebook Histogram Distance} (CHD), a training-free distribution metric in token space, and \emph{Code Mixture Model Score} (CMMS), a no-reference quality metric learned from synthetic degradations of token sequences. To stress-test metrics under broad distribution shifts, we further propose \emph{VisForm}, a benchmark of 210K images spanning 62 visual forms and 12 generative models with expert annotations. Across AGIQA, HPDv2/3, and VisForm, our token-based metrics achieve state-of-the-art correlation with human judgments. We will release all code and datasets to facilitate future research, with the code publicly available at \url{https://github.com/zexiJia/1d-Distance}.
\end{abstract}    
\section{Introduction}

\begin{figure*}[h!]
\begin{center}
\includegraphics[width=0.8\linewidth]{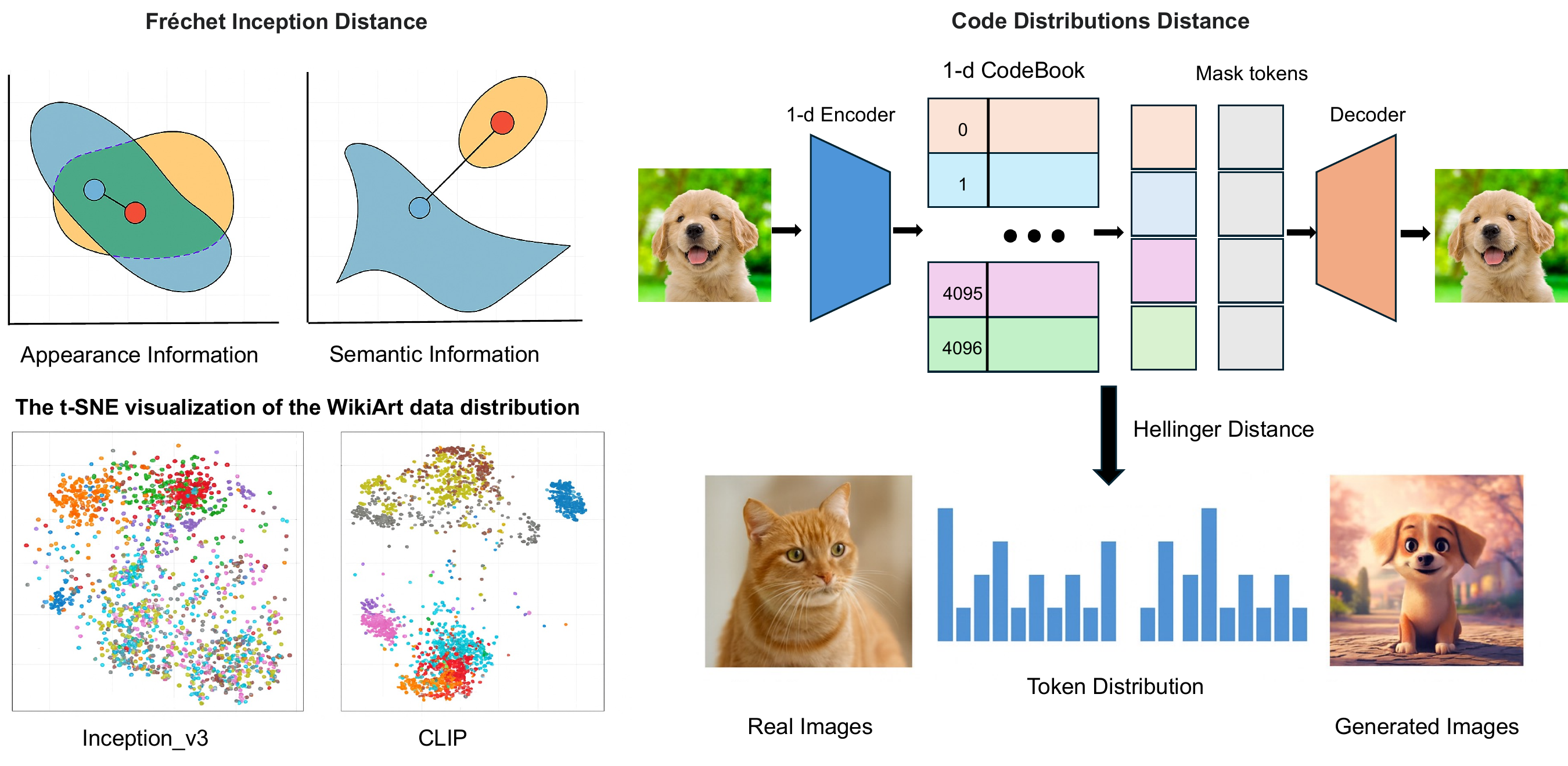}
\end{center}
\caption{
\textbf{From feature distributions to token statistics.}
Conventional metrics such as Fréchet Inception Distance (FID) operate on continuous semantic features and assume a Gaussian distribution in feature space (left), which makes them insensitive to appearance details (e.g., texture, style) and unreliable on non-Gaussian data such as artistic or medical images. Our approach (right) quantizes images into a discrete vocabulary of 1D tokens and compares empirical token statistics directly.
}
\label{fig:intro}
\end{figure*}

The rapid progress of generative models, from GANs~\cite{goodfellow2014generative} to diffusion models~\cite{ho2020denoising,rombach2022high}, has enabled high-quality image synthesis across many domains. In contrast, evaluation still relies heavily on feature-distribution metrics such as FID~\cite{heusel2017gans}, which can disagree with human perception~\cite{stein2023exposing}. By summarizing complex image distributions as Gaussians over continuous recognition features, these metrics underweight fine-grained artifacts, local compositional failures, and many aspects of visual quality, making model comparison and debugging difficult.

Most recent work improves evaluation along two lines. The first line modifies the feature space or the distributional assumption, for example by adopting CLIP~\cite{radford2021learning} or DINO~\cite{stein2023exposing} features, or by using kernel MMD instead of Gaussian Fréchet distance~\cite{jayasumana2024rethinking}. However, all such methods compress each image into a single feature vector, discarding spatial structure and local coherence signals that are crucial for detecting artifacts. The second line trains learned metrics directly on human preference data~\cite{wu2023human,kirstain2023pick,xu2023imagereward}, which improves alignment but requires large-scale annotations and often exhibits domain shift when applied to new styles.

We argue that this tension stems from a shared design choice: evaluating generative models in the space of continuous recognition features (Figure~\ref{fig:intro}). Instead, we propose to work in the space of \emph{discrete} visual tokens. Modern tokenizers such as TiTok~\cite{yu2024an} learn a rich visual vocabulary and quantize an image into a compact sequence of codebook indices. Discrete codebooks and their histograms have been widely used for compression and as internal representations in generative models, and occasionally as heuristic signals (e.g., for anomaly detection or overfitting analysis). Our goal is different: we treat the token space itself as a \emph{primary evaluation domain}, and systematically develop and study metrics that operate purely on token statistics. To cover diverse visual domains, we retrain TiTok on a large, heterogeneous image collection, obtaining a 1D tokenizer that captures both semantic content and perceptual details. Our central hypothesis is that statistics over this discrete vocabulary provide a more faithful and interpretable basis for evaluation: token frequencies and co-occurrences directly reflect what structures a model generates, without imposing Gaussian assumptions or collapsing spatial information.

Building on this view, we introduce two complementary metrics. \textbf{Codebook Histogram Distance (CHD)} measures distribution fidelity by computing unigram and local co-occurrence histograms over token sequences and evaluating a Hellinger distance between real and generated sets. This training-free metric compares visual “vocabulary” usage and local “grammar”, making it sensitive to both semantic shifts and stylistic changes. \textbf{Code Mixture Model Score (CMMS)} assesses single-image quality using a lightweight regressor on token sequences that is learned but self-supervised: we construct a synthetic degradation model in token and pixel space that injects uniform tokens and common distortions, and train CMMS to map the resulting token patterns to a continuous quality score. CMMS is therefore a learned metric, but it does not rely on human preference labels for training, and instead exploits automatically generated corruptions as supervision.

To stress-test metrics under broad distribution shifts, we further introduce \textbf{VisForm}, a benchmark of 210K images spanning 62 visual forms (e.g., photographs, artistic styles, 3D renders, scientific diagrams) and 12 generative models. Each image is annotated by experts along 14 perceptual dimensions, providing a rich testbed for analyzing metric–human alignment across models and domains.

Our contributions are three-fold:

\begin{itemize}
\item We propose a discrete-token paradigm for generative model evaluation, shifting from continuous recognition features to structured codebook statistics as a first-class evaluation space.
\item We introduce two token-space metrics: CHD, a training-free distribution metric, and CMMS, a no-reference quality metric on token sequences, both showing strong alignment with human judgments across multiple benchmarks.
\item We present VisForm, a large-scale benchmark covering 62 diverse visual forms with expert annotations, enabling comprehensive cross-domain evaluation of generative models and quality metrics. We will release all code, models, and data to facilitate future research.
\end{itemize}

\section{Related Work}
The evaluation of generative models is crucial for their advancement, yet finding a reliable metric that captures both distributional fidelity and perceptual quality remains a challenge. Existing approaches often face inherent trade-offs.

\textbf{Non-reference Metrics:}
Non-reference metrics are widely used for open-ended generation tasks as they do not require ground truth images. Distribution-based methods, such as the Inception Score (IS)~\cite{salimans2016improved} and Fr\'echet Inception Distance (FID)~\cite{heusel2017gans}, compare feature distributions of real and generated samples. Despite its widespread adoption, FID's Gaussian assumption and reliance on Inception features can limit its reliability across diverse image distributions~\cite{jayasumana2024rethinking,stein2023exposing}. While variants like CLIP-FID~\cite{kynkaanniemi2022role} and DINO-FID~\cite{stein2023exposing} improve the feature encoder, they inherit the same structural weaknesses. Alternatively, single-image approaches, including NIQE~\cite{mittal2012making} and MUSIQ~\cite{ke2021musiq}, aim to assess perceptual quality on a per-image basis. However, these methods often fail to capture subtle artifacts and demonstrate limited robustness across diverse visual content.

\textbf{Reference-based Metrics:}
When a reference image is available, classical measures such as PSNR and SSIM assess pixel-level fidelity, while LPIPS~\cite{zhang2018unreasonable} evaluates learned perceptual features. These methods are effective for tasks like image restoration but are unsuitable for generative settings where multiple outputs can be equally valid. For text-to-image synthesis, semantic alignment is often evaluated using metrics like CLIP-Score~\cite{hessel2021clipscore}. While useful for measuring prompt consistency, these methods do not directly account for visual quality and may reward images that are textually aligned but perceptually flawed.

\textbf{Human Preference Modeling:}
To bridge the gap between automated metrics and human perception, a recent paradigm learns quality scores directly from user annotations. Models like HPS~\cite{wu2023human} and PickScore~\cite{kirstain2023pick} demonstrate a strong correlation with subjective judgments. Newer approaches such as Q-Align~\cite{qalign2024icml} and DeQA-Score~\cite{wu2025deqa} further refine these architectures. However, these methods are not without limitations; they require large-scale, costly annotations, are prone to dataset biases, and generalize poorly to unseen distributions.

In contrast, our approach leverages discrete token distributions to provide a scalable, reference-free, and perceptually aligned evaluation without relying on parametric assumptions or costly human supervision.

\begin{figure*}[h!]
\begin{center}
\includegraphics[width=\linewidth]{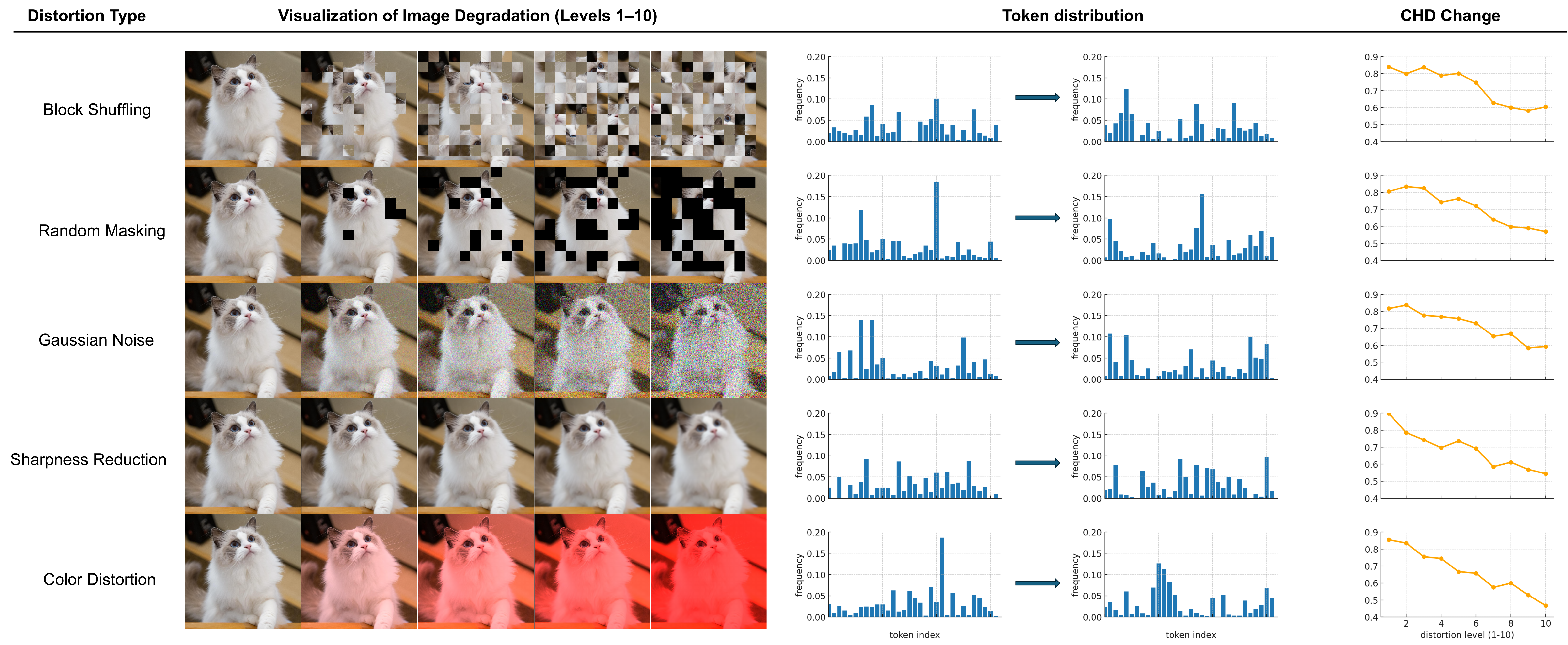}
\end{center}
\caption{
\textbf{Sensitivity of Token Distributions to Image Degradation. }To demonstrate how our discrete token space captures perceptual degradations, we apply 10 levels of progressive distortion to a set of 1,000 images and analyze the resulting shifts in their token distributions. As the severity of distortions like Gaussian noise or block shuffling increases (left), a small subset of perceptually-sensitive tokens exhibits consistent and predictable shifts in their distribution (middle). Our Codebook Histogram Distance (CHD) effectively aggregates these subtle changes, showing a robust, monotonic increase with the degradation level across all distortion types (right).
}
\label{fig:method}
\end{figure*}
\section{Analysis}

\subsection{Limitations of Feature-Distribution Metrics}

Distribution-based metrics such as FID are widely used for evaluating generative models, but they suffer from a fundamental objective mismatch: features trained for recognition are optimized to be invariant to appearance variations (texture, sharpness, local coherence) that humans are explicitly sensitive to.

FID computes the Fréchet distance between Gaussians fitted to Inception-V3 features:
\begin{equation}
\text{FID} = \|\mu_r - \mu_g\|_2^2 + \mathrm{Tr}\!\left(\Sigma_r + \Sigma_g - 2(\Sigma_r\Sigma_g)^{1/2}\right),
\end{equation}
where $(\mu_r,\Sigma_r)$ and $(\mu_g,\Sigma_g)$ are the empirical means and covariances of real and generated features. In practice, real and generated features are often multi-modal and skewed rather than Gaussian, making the Fréchet approximation inaccurate. The covariance estimates are also noisy in high dimensions, and the matrix square root is numerically unstable, leading to sensitivity to sample size and implementation details.

From an information-theoretic perspective, decomposing an image into semantic content $x_s$ and appearance $x_a$, and letting $\phi$ be an encoder, the chain rule gives
\begin{equation}
I(x_s,x_a;\phi(x)) = I(x_s;\phi(x)) + I(x_a;\phi(x)\mid x_s).
\end{equation}
Recognition training explicitly increases $I(x_s;\phi(x))$ while encouraging invariance to appearance, thereby reducing $I(x_a;\phi(x)\mid x_s)$ and discarding quality-relevant cues. For a Markov chain $q \to x \to \phi(x)$, where latent quality $q$ influences the image $x$ which is then encoded, the data processing inequality implies
\begin{equation}
I(q;x) \;\ge\; I\bigl(q;\phi(x)\bigr),
\end{equation}
so any compression that is not explicitly optimized for quality must lose information about $q$.

Global pooling further weakens sensitivity to spatial structure. Most encoders apply spatial averaging $\phi(x) = \tfrac{1}{HW}\sum_{i,j} f_{i,j}(x)$ over feature maps $f_{i,j}(x)$, collapsing local arrangements into global summary statistics and reducing sensitivity to localized artifacts.

Recent variants partially address these issues. CLIP-FID~\cite{kynkaanniemi2022role} and DINO-FID~\cite{stein2023exposing} replace Inception features with CLIP or DINO, but inherit the same architectural constraints (global pooling, semantic invariance). CMMD~\cite{jayasumana2024rethinking} replaces the Gaussian assumption with a kernel maximum mean discrepancy,
\begin{equation}
\mathrm{MMD}_k^2 = \mathbb{E}_{x,x'}[k(x,x')] + \mathbb{E}_{y,y'}[k(y,y')] - 2\mathbb{E}_{x,y}[k(x,y)],
\end{equation}
where $k$ is a kernel. This removes parametric assumptions but introduces a new sensitivity: in high dimensions, the statistical power of MMD depends critically on the kernel bandwidth, and poorly tuned kernels require substantially more samples to detect distributional shifts.

\subsection{Discrete Codes as a Foundation for Evaluation}

Continuous features are shaped by objectives that encourage invariance and compression. In contrast, discrete tokenizations are trained to reconstruct images and thus naturally retain both semantic content and appearance details in a unified, lossless index space. Modern tokenizers are highly compact: TiTok~\cite{yu2024an} reconstructs $256\times256$ images from as few as 32 tokens with high perceptual fidelity, and empirical analysis~\cite{lao2025highly} shows individual token positions encode disentangled attributes such as blur, lighting, and sharpness.

Conceptually, classification features learn invariances to appearance, whereas discrete codes learn \emph{equivariant} representations that change predictably with both content and style. Writing the token sequence as $\mathbf{c}=[c_1,\ldots,c_N]$, and $x=(x_s,x_a)$ as semantic and appearance components, we have
\begin{equation}
I(x;\mathbf{c}) = I(x_s;\mathbf{c}) + I(x_a;\mathbf{c}) + I(x_s,x_a;\mathbf{c}),
\end{equation}
where the interaction term captures how content and appearance are jointly encoded. In practice, the tokenizer is trained so that $\mathbf{c}$ retains sufficient information for reconstruction, rather than enforcing invariance.

Discrete codes also make distributional analysis tractable. Given a codebook $\mathcal{V}$ of size $K$, we can factorize the joint distribution as $p(\mathbf{c}) = \prod_{i=1}^N p(c_i \mid c_{<i})$, preserving rich dependencies that global pooling erases. Quality naturally manifests in these statistics: natural images produce highly structured, low-entropy token patterns, while degraded images produce more random, high-entropy ones, i.e.,
\begin{equation}
H(\mathbf{c} \mid q_{\text{high}}) \;<\; H(\mathbf{c} \mid q_{\text{low}}).
\end{equation}
Similarly, spatial coherence can be quantified through mutual information between adjacent tokens,
\begin{equation}
I(c_i;c_{i+1}) = H(c_i) + H(c_{i+1}) - H(c_i,c_{i+1}),
\end{equation}
which decreases when artifacts disrupt natural co-occurrence patterns. In practice, the learned codebook transforms quality assessment from a high-dimensional continuous problem into counting and comparing token statistics: frequent patterns correspond to natural structures, while rare or inconsistent combinations act as reliable signals of artifacts.

\section{Method}

\subsection{Codebook Histogram Distance}
\label{sec:chd}

To measure distributional discrepancy between sets of real and generated images, we propose \emph{Codebook Histogram Distance} (CHD). We first discretize each $256 \times 256$ image using a pre-trained TiTok encoder~\cite{yu2024an}, which maps the image to a sequence of $N=128$ discrete tokens from a codebook $\mathcal{V}$ with $|\mathcal{V}| = 4096$. This unified 1D tokenization allows us to compare distributions in a non-parametric way, avoiding Gaussian assumptions and feature learning.

\textbf{Unigram statistics (CHD-1D).}
For a set of images $\mathcal{S}$, we compute the empirical unigram histogram
\begin{equation}
h_\mathcal{S}^{(1)}(v) \;=\; \frac{1}{|\mathcal{S}| \cdot N} \sum_{I \in \mathcal{S}} \sum_{i=1}^N \mathbb{I}[c_i(I) = v], \quad v \in \mathcal{V},
\end{equation}
where $c_i(I)$ is the $i$-th token of image $I$. The 1D CHD between real images $\mathcal{R}$ and generated images $\mathcal{G}$ is the Hellinger distance between their histograms:
\begin{equation}
\text{CHD-1D}(\mathcal{R}, \mathcal{G}) \;=\; \frac{1}{\sqrt{2}} \bigl\| \sqrt{h_\mathcal{R}^{(1)}} - \sqrt{h_\mathcal{G}^{(1)}} \bigr\|_2 \;\in [0,1].
\end{equation}
CHD-1D efficiently measures whether a model learns the correct \emph{visual vocabulary}.

\textbf{Spatial co-occurrence statistics (CHD-2D).}
One-dimensional adjacency along the token sequence imposes an artificial order that does not align with the underlying image grid. To capture local structure, we introduce a second-order statistic based on 2D spatial adjacency.

We view the quantized image as a token grid $\{c(\mathbf{p})\}_{\mathbf{p} \in \Omega_I}$, with pixel positions $\mathbf{p} = (x,y)$. We define a small set of displacement vectors $\mathcal{D}$ (e.g., $\mathcal{D}=\{(1,0),(0,1)\}$ for rightward and downward neighbors, avoiding double-counting). For each $\Delta \in \mathcal{D}$, we compute a directed co-occurrence distribution
\begin{equation}
h_\mathcal{S}^{(2)}(u,v;\Delta) \;=\; \frac{1}{Z_{\mathcal{S},\Delta}} \sum_{I\in\mathcal{S}} \sum_{\substack{\mathbf{p} \in \Omega_I \\ \mathbf{p}+\Delta \in \Omega_I}} \mathbb{I}[c(\mathbf{p})=u,\, c(\mathbf{p}+\Delta)=v],
\end{equation}
where $Z_{\mathcal{S},\Delta}$ is the total number of valid adjacent pairs for normalization. To remove the ordering within a pair, we symmetrize
\begin{equation}
\tilde{h}_\mathcal{S}^{(2)}(u,v;\Delta) \;=\; \tfrac{1}{2}\bigl(h_\mathcal{S}^{(2)}(u,v;\Delta) + h_\mathcal{S}^{(2)}(v,u;\Delta)\bigr),
\end{equation}
and average over $\Delta$ to obtain an orientation-robust co-occurrence:
\begin{equation}
\bar{h}_\mathcal{S}^{(2)}(u,v) \;=\; \frac{1}{|\mathcal{D}|} \sum_{\Delta \in \mathcal{D}} \tilde{h}_\mathcal{S}^{(2)}(u,v;\Delta).
\end{equation}
We only store entries $(u,v)$ that appear at least once in $\mathcal{S}$, yielding a sparse representation of $\bar{h}_\mathcal{S}^{(2)}$ in practice. The 2D CHD is again a Hellinger distance:
\begin{equation}
\text{CHD-2D}(\mathcal{R},\mathcal{G}) \;=\; \frac{1}{\sqrt{2}} \bigl\| \sqrt{\mathrm{vec}(\bar{h}_\mathcal{R}^{(2)})} - \sqrt{\mathrm{vec}(\bar{h}_\mathcal{G}^{(2)})} \bigr\|_2 \;,
\end{equation}
where $\mathrm{vec}(\cdot)$ flattens the sparse co-occurrence matrix into a vector over observed pairs.

CHD-1D measures whether the model matches the \emph{composition} of visual tokens, while CHD-2D measures whether these tokens are combined with the correct local \emph{grammar}. We define the final CHD metric as their arithmetic mean:
\begin{equation}
\text{CHD}(\mathcal{R},\mathcal{G}) \;=\; \tfrac{1}{2}\bigl(\text{CHD-1D}(\mathcal{R},\mathcal{G}) + \text{CHD-2D}(\mathcal{R},\mathcal{G})\bigr).
\end{equation}
This composite score provides a balanced, training-free assessment of both global vocabulary fidelity and local structural coherence.

\subsection{Code Mixture Model Score}
\label{sec:cmms}

We next introduce \emph{Code Mixture Model Score} (CMMS), a no-reference image quality metric that operates directly on discrete token sequences. CMMS is trained to regress a quality score from tokenized images, using a synthetic degradation model that mimics common generative artifacts.

\textbf{Token corruption.}
Given a token sequence $\{c_i\}_{i=1}^N$, we first apply an independent corruption process
\begin{equation}
\tilde{c}_i \sim 
\begin{cases}
c_i & \text{with probability } 1-p, \\
\mathcal{U}(\mathcal{V}) & \text{with probability } p,
\end{cases}
\label{eq:cmms_mixture}
\end{equation}
where $\mathcal{U}(\mathcal{V})$ is the uniform distribution over the codebook. This uniform token injection simulates unpredictable local artifacts (e.g., spurious patterns or texture glitches) frequently observed in generative outputs.

\textbf{Semantic fragment swapping and pixel-space degradation.}
Uniform corruption alone does not capture higher-level structural failures. We therefore introduce two additional degradations:

\begin{itemize}[leftmargin=*, noitemsep, topsep=2pt]
    \item \emph{Semantic fragment swapping.} We exchange spatially contiguous token blocks between images or between distant regions of the same image, simulating object-level inconsistencies such as misplaced parts, broken limbs, or repeated fragments.
    \item \emph{Pixel-space augmentation.} Before tokenization, we apply a set of standard distortions: Gaussian blur ($\sigma \in [0.5, 3.0]$), JPEG compression (quality in $[10, 90]$), Gaussian noise ($\sigma \in [0.01, 0.1]$), random occlusion (covering 10\%–40\% of the area), and photometric changes (sharpening, contrast, brightness, saturation). These operations emulate low-level degradations such as over/under-sharpening, compression artifacts, and abnormal exposure.
\end{itemize}

Together, these mechanisms generate a rich family of degraded token sequences that resemble both local noise and high-level structural errors in generative models.

\textbf{Quality mapping and regressor.}
We associate each degraded sample with a target quality score determined by the corruption severity $p$:
\begin{equation}
q(p) \;=\; \exp(-20p), \qquad p \in [0,0.3].
\end{equation}
This exponential mapping reflects the non-linear sensitivity of human vision: small perturbations at high quality lead to noticeable drops, while additional degradation at already low quality has a smaller perceived effect. We choose the constant 20 via a hyperparameter search on a held-out validation set, maximizing Spearman correlation between predicted scores and human ratings (Table~\ref{tab:ablation_study}).

The score regressor takes the $N$ tokens as input, embeds them into a 512-dimensional space with sinusoidal positional encodings, and passes the sequence through a 2-layer Transformer encoder with 8 attention heads per layer. Global average pooling yields a representation $\mathbf{g} \in \mathbb{R}^{512}$, which a 2-layer MLP maps to a scalar prediction $\hat{q} \in [0,1]$. We train CMMS on tokenized ImageNet-1K images only, and use it \emph{without further fine-tuning} on all downstream datasets and on VisForm.

\begin{figure}[t!]
\begin{center}
\includegraphics[width=\linewidth]{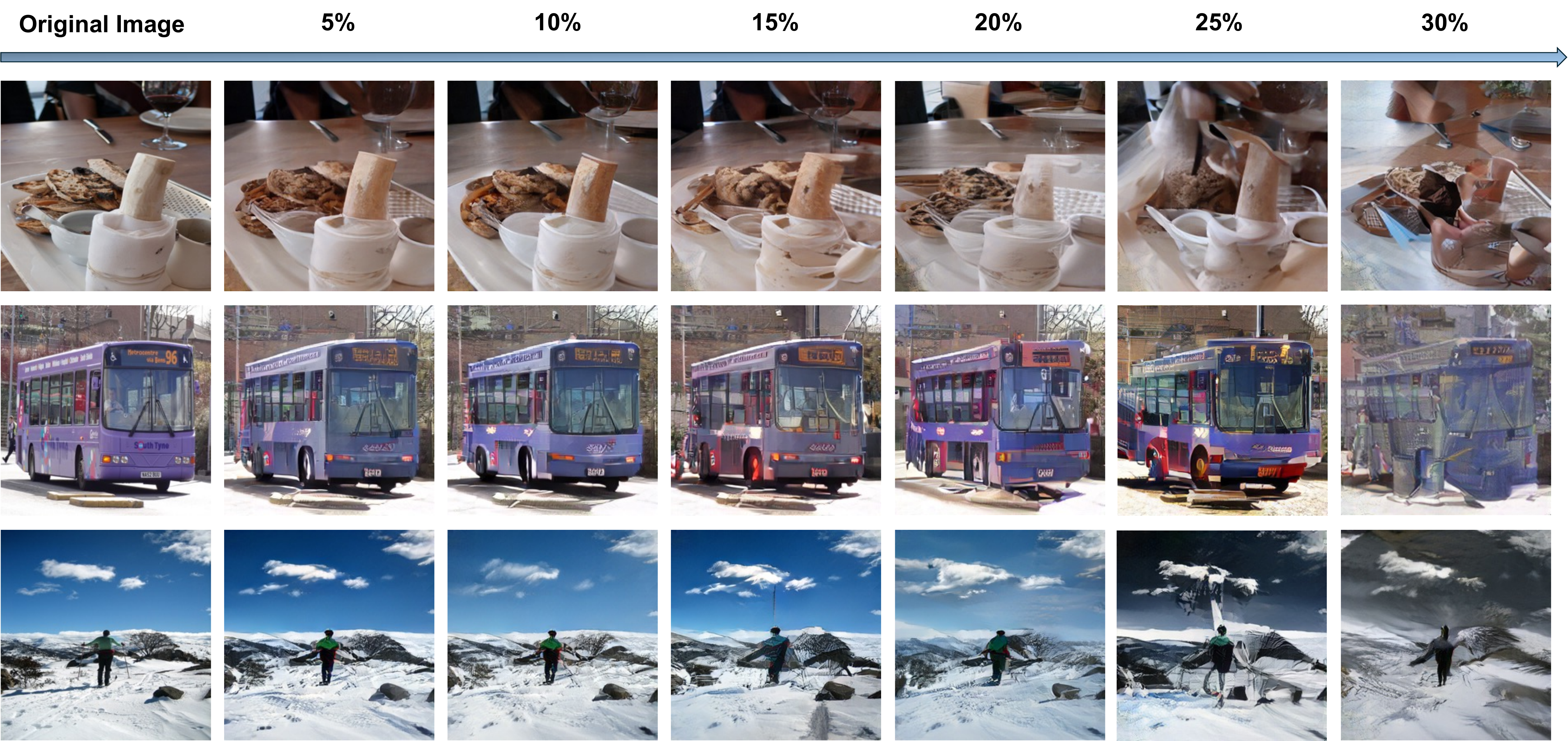}
\end{center}
\caption{\textbf{Code Mixture Model Degradation.} CMMS is trained on token sequences obtained from natural images that are progressively corrupted via uniform token injection, semantic fragment swapping, and pixel-space distortions, without any human labels.}
\label{fig:enhance}
\end{figure}

\subsection{The VisForm Benchmark}
\label{sec:visform}

Existing benchmarks for generative model evaluation predominantly target natural images or narrow domains, limiting our ability to study quality metrics under broad distribution shifts. We therefore introduce \emph{VisForm}, a large-scale benchmark of 210,000 images spanning 62 visual domains and 12 generative models.

\textbf{Domains and models.}
VisForm covers a wide spectrum of visual forms, including but not limited to photorealistic portraits, landscapes, product photos, watercolor and oil paintings, anime and comics, 3D renders, medical imagery, scientific diagrams, and UI/infographics. Images are generated by 12 representative models covering different architectures and training recipes (e.g., diffusion, consistency models, and autoregressive transformers). Each sample is labeled by its visual domain and source model, enabling analysis along two axes: domain-specific behavior and model-specific characteristics.

\textbf{Perceptual annotations.}
Each image is annotated along 14 perceptual dimensions such as overall quality, composition, semantic coherence, color harmony, lighting realism, texture naturalness, artifact severity, and text rendering quality. Every image receives ratings from at least three independent expert annotators. We enforce quality control through calibration rounds and outlier filtering, achieving inter-annotator agreement of Kendall's $W > 0.75$. Final scores per dimension are obtained by averaging ratings after majority filtering.

\textbf{Usage and availability.}
VisForm is used \emph{exclusively} for evaluating quality metrics and generative models in our experiments; CMMS is never trained or fine-tuned on VisForm. Detailed domain taxonomy, model configurations, annotation protocols, and dataset statistics are provided in the supplementary material. This design makes VisForm a challenging and diverse testbed for assessing how well quality metrics generalize across visual domains, generative architectures, and perceptual factors.

\section{Experiments}

\begin{table*}[h!]
\begin{center}
\caption{Evaluation of different generative models on AGIQA~\cite{agiqa3k}.}
\renewcommand{\arraystretch}{0.8} 
\setlength{\tabcolsep}{2.5pt}
\begin{tabular}{c|c|cccccc|ccc}
    \toprule
    Methods & Reference & \multicolumn{6}{c|}{AGIQA} & Spearman$\uparrow$ & Kendall$\uparrow$ & N-MSE$\downarrow$ \\
    \cmidrule(lr){3-8}
     & & AttnGAN & DALLE2 & Glide & Midjourney & SD-1.5 & SD-XL &  &  &  \\
    \midrule
    Human$\uparrow$ & -- & 0.986 & 2.624 & 1.092 & 3.007 & 2.752 & 3.298 & -- & -- & -- \\
    \midrule
    FID$\downarrow$\cite{heusel2017gans} & NeurIPS’17 & 77.7 & 77.5 & 101.45 & 59.45 & 41.2 & 78.45 & 0.771 & 0.600 & 0.119 \\
    KID$\downarrow$\cite{binkowski2018demystifying} & ICLR’18 & 0.031 & 0.024 & 0.076 & 0.033 & 0.025 & 0.036 & 0.486 & 0.333 & 0.236 \\
    IS$\uparrow$\cite{salimans2016improved} & NeurIPS’16 & 13.8 & 15.8 & 16.8 & 20.8 & 26.6 & 15.2 & 0.543 & 0.467 & 0.224 \\
    CLIP-FID$\downarrow$\cite{kynkaanniemi2022role} & ICLR’23 & 0.607 & 0.547 & 0.676 & 0.572 & 0.451 & 0.656 & 0.714 & 0.467 & 0.170 \\
    DINO-FID$\downarrow$\cite{stein2023exposing} & NeurIPS’23 & 333.3 & 340.0 & 764.9 & 413.6 & 170.6 & 316.6 & 0.657 & 0.600 & 0.135 \\
    CMMD$\downarrow$\cite{jayasumana2024rethinking} & CVPR’24 & 0.180 & 0.097 & 0.156 & 0.111 & 0.098 & 0.105 & 0.657 & 0.600 & 0.142 \\
    \rowcolor{blue!10}CHD$\downarrow$ & Ours & 0.135 & 0.128 & 0.162 & 0.099 & 0.131 & 0.134 & \textbf{0.829} & \textbf{0.733} & \textbf{0.112} \\
    \midrule
    MUSIQ$\uparrow$\cite{ke2021musiq} & ICCV’21 & 46.6 & 55.1 & 36.8 & 51.7 & 60.2 & 66.8 & 0.486 & 0.333 & 0.342 \\
    CLIP-IQA$\uparrow$\cite{wu2023clipiqa} & AAAI’23 & 0.772 & 0.785 & 0.812 & 0.765 & 0.761 & 0.780 & -0.086 & -0.067 & 0.604 \\
    QualiCLIP$\uparrow$\cite{quali2024arxiv} & arXiv’24 & 0.511 & 0.635 & 0.374 & 0.550 & 0.641 & 0.713 & 0.771 & 0.733 & 0.122 \\
    DeQA-Score$\uparrow$\cite{wu2025deqa} & CVPR’25 & 2.117 & 3.139 & 1.642 & 3.169 & 3.665 & 4.014 & 0.886 & 0.733 & 0.118 \\
    \rowcolor{blue!10}CMMS$\uparrow$ & Ours & 0.570 & 0.588 & 0.512 & 0.595 & 0.592 & 0.620 & \textbf{0.943} & \textbf{0.867} & \textbf{0.050} \\
    \bottomrule
\end{tabular}
\label{tab:performance_comparison4}
\end{center}
\end{table*}

\begin{table*}[h!]
\centering
\renewcommand{\arraystretch}{0.8}
\caption{Evaluation of different generative models on HPDv3~\cite{ma2025hpsv3widespectrumhumanpreference}.}
\setlength{\tabcolsep}{1.4pt}
\begin{tabular}{l|cccccccccc|ccc}
\toprule
Metric & Real & Kolors & Flux & Infinity & SD-XL & Hunyuan & SD-3 & SD-2.0 & SD-1.4 & Glide & Spearman$\uparrow$ & Kendall$\uparrow$ & N-MSE$\downarrow$ \\
\midrule
Human$\uparrow$        & 11.48 & 10.55 & 10.43 & 10.26 &  8.20 &  8.19 &  5.31 & -0.24 & -3.27 & -7.46 & --    & --    & --    \\
\midrule
FID$\downarrow$\cite{heusel2017gans}      & 24.7  & 41.2  & 35.3  & 36.8  & 35.7  & 35.7  & 30.5  & 53.9  & 41.6  & 64.1  & 0.648 & 0.467 & 0.043 \\
IS$\uparrow$\cite{salimans2016improved}   & 26.0  & 27.5  & 30.1  & 27.0  & 29.9  & 22.5  & 32.3  & 13.7  & 24.7  & 20.3  & 0.491 & 0.289 & 0.085 \\
KID$\downarrow$\cite{binkowski2018demystifying}      & 0.010 & 0.022 & 0.018 & 0.020 & 0.019 & 0.017 & 0.015 & 0.027 & 0.021 & 0.042 & 0.515 & 0.333 & 0.045 \\
CLIP-FID$\downarrow$\cite{kynkaanniemi2022role} & 0.264 & 0.328 & 0.276 & 0.299 & 0.306 & 0.297 & 0.253 & 0.385 & 0.328 & 0.447 & 0.491 & 0.378 & 0.043 \\
DINO-FID$\downarrow$\cite{stein2023exposing} & 171.0 & 216.4 & 196.9 & 160.3 & 328.1 & 282.3 & 268.9 & 544.7 & 290.0 & 527.7 & 0.782 & 0.556 & 0.045 \\
CMMD$\downarrow$\cite{jayasumana2024rethinking}     & 0.050 & 0.070 & 0.054 & 0.056 & 0.060 & 0.060 & 0.049 & 0.093 & 0.065 & 0.114 & 0.527 & 0.467 & 0.048 \\
\rowcolor{blue!10}{CHD (Ours)$\downarrow$} & 0.036 & 0.049 & 0.040 & 0.046 & 0.053 & 0.064 & 0.047 & 0.066 & 0.087 & 0.089 & \textbf{0.867} & \textbf{0.778} & \textbf{0.017} \\
\midrule
MUSIQ$\uparrow$\cite{ke2021musiq}      & 61.1  & 68.0  & 65.1  & 65.6  & 63.7  & 63.6  & 65.7  & 59.6  & 62.4  & 35.7  & 0.503 & 0.422 & 0.061 \\
CLIP-IQA$\uparrow$\cite{wu2023clipiqa}   & 0.755 & 0.772 & 0.762 & 0.758 & 0.760 & 0.788 & 0.791 & 0.779 & 0.769 & 0.786 & 0.612 & 0.422 & 0.399 \\
QualiCLIP$\uparrow$\cite{quali2024arxiv}      & 0.694 & 0.753 & 0.721 & 0.730 & 0.724 & 0.703 & 0.724 & 0.618 & 0.703 & 0.407 & 0.503 & 0.422 & 0.055 \\
DeQA-Score$\uparrow$\cite{wu2025deqa}      & 4.232 & 4.317 & 4.228 & 4.193 & 4.076 & 3.862 & 4.216 & 2.819 & 3.671 & 1.737 & 0.836 & 0.689 & 0.026 \\
\rowcolor{blue!10}{CMMS (Ours)$\uparrow$}  & 0.629 & 0.618 & 0.609 & 0.612 & 0.609 & 0.608 & 0.606 & 0.589 & 0.587 & 0.529 & \textbf{0.872} & \textbf{0.778} & \textbf{0.018} \\
\bottomrule
\end{tabular}
\label{tab:generation_metrics}
\end{table*}

\begin{figure*}[h!]
\begin{center}
\includegraphics[width=0.8\linewidth]{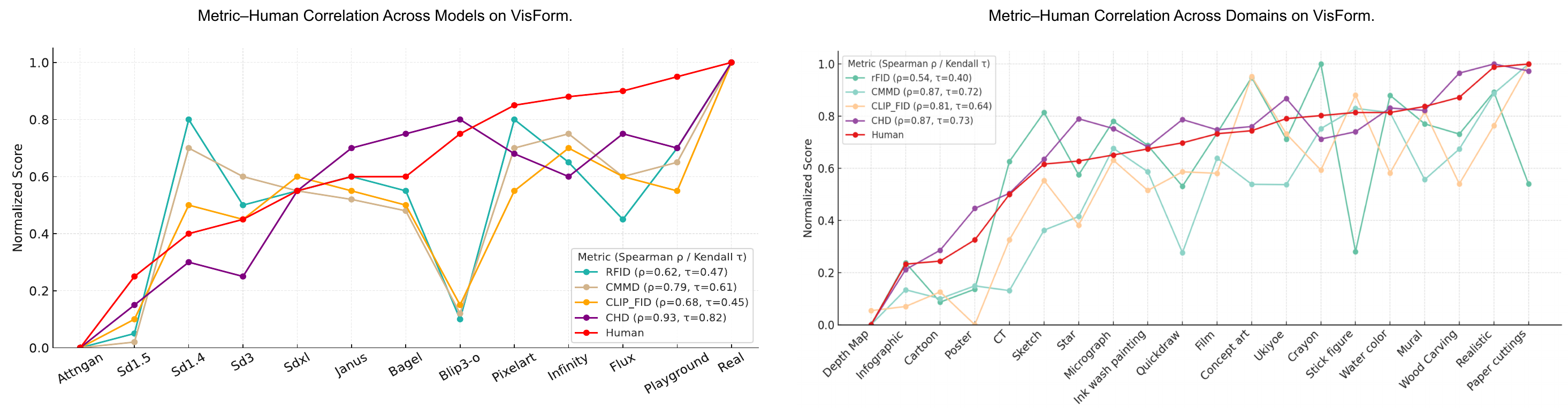}
\end{center}
\caption{
\textbf{Metric–human correlation on VisForm across models and domains.} All metrics are normalized to $[0,1]$, higher is better.
}
\label{fig:visform}
\end{figure*}

\subsection{Datasets and Evaluation Protocol}

\textbf{Datasets.}
We evaluate on three human preference benchmarks and one large-scale natural image dataset.  
AGIQA~\cite{agiqa3k} contains 2{,}982 AI-generated images from GAN, autoregressive, and diffusion models with 125{,}244 human ratings.  
HPDv2~\cite{hpsv2} includes 430{,}060 images and 798{,}090 ratings, while HPDv3~\cite{ma2025hpsv3widespectrumhumanpreference} extends this to 1.08M text–image pairs and 1.17M human comparisons, covering ten additional models and real-world Midjourney user preferences.  
CMMS is trained once on the 1.28M images of ImageNet-1K and applied to all benchmarks without fine-tuning.

\textbf{Metrics.}
We quantify agreement between objective metrics and human judgments using Spearman’s rank correlation, Kendall’s tau, and normalized mean squared error (N-MSE). Spearman and Kendall measure rank- and pairwise-level consistency, respectively, while N-MSE captures normalized deviation between predicted scores and human ratings. For preference prediction, we additionally report pairwise accuracy: the fraction of image pairs for which the metric selects the same winner as humans.

\subsection{Implementation Details}

We retrain the TiTok encoder on 100M images from DataComp~\cite{datacomp} to better cover diverse visual domains, following the official setup~\cite{yu2024an}. Training takes 214 hours on 8 NVIDIA A100 GPUs. All experiments in this paper use this retrained tokenizer.

Our implementation is in PyTorch. At inference time, TiTok supports batch sizes up to 1024 on a single A100 GPU. CHD only requires accumulating token histograms, adding negligible overhead beyond encoding. CMMS uses a lightweight Transformer–MLP regressor and can process up to 2{,}048 token sequences per batch, achieving over 1{,}000 images per second. Training CMMS for 200 epochs on ImageNet takes less than 24 hours using AdamW with learning rate $1\times10^{-4}$, batch size 512, and weight decay 0.01.

\begin{table}[t]
\centering
\renewcommand{\arraystretch}{1.0}
\setlength{\tabcolsep}{6pt}
\caption{Preference prediction on human preference benchmarks.}
\resizebox{\columnwidth}{!}{
\begin{tabular}{l|cccc}
\toprule
\textbf{Preference Model} & AGIQA & HPDv2 & HPDv3 & VisForm \\
\midrule
CLIPScore\cite{hessel2021clipscore} & 63.4 & 65.1 & 48.6 & 58.2 \\ 
MUSIQ\cite{ke2021musiq} & 51.3 & 52.8 & 39.4 & 47.2 \\ 
CLIP-IQA\cite{wu2023clipiqa} & 63.8 & 65.5 & 48.9 & 58.6 \\
QualiCLIP\cite{quali2024arxiv} & 66.7 & 68.6 & 51.2 & 61.3 \\ 
MDIQA\cite{Yao_2025_MDIQA} & 66.3 & 70.1 & 51.1 & 64.5 \\ 
DeQA-Score\cite{wu2025deqa} & 68.7 & 70.6 & 52.7 & 63.1 \\ 
\midrule 
\rowcolor{blue!10}CMMS (Ours) & \textbf{71.5} & \textbf{74.9} & \textbf{61.3} & \textbf{66.7} \\
\bottomrule
\end{tabular}}
\label{tab:pref_models_datasets}
\end{table}

\begin{table}[t]
\centering
\caption{Ablation study of CHD (N-MSE$\downarrow$) and CMMS (Acc$\uparrow$).}
\setlength{\tabcolsep}{3pt}
\renewcommand{\arraystretch}{0.9}
\resizebox{\columnwidth}{!}{%
\begin{tabular}{l|cccc}
\toprule
\textbf{Setting} & \textbf{AGIQA} & \textbf{HPDv2} & \textbf{HPDv3} & \textbf{VisForm} \\
\midrule
\multicolumn{5}{c}{\textit{CHD / N-MSE $\downarrow$}} \\
\midrule
\multicolumn{5}{l}{\textit{Tokenizer Architecture}} \\
VQ-VAE (2D tokens) \cite{vandenOord2017neural} & 0.268 & 0.152 & 0.114 & 0.147 \\
VQGAN (2D tokens) \cite{esser2021taming}       & 0.245 & 0.145 & 0.123 & 0.139 \\
Instella-T2I (1D tokens) \cite{instella-t2i}   & 0.118 & \textbf{0.028} & 0.019 & 0.027 \\
\rowcolor{blue!10} TiTok (1D tokens) \cite{yu2024an} & \textbf{0.112} & 0.030 & \textbf{0.017} & \textbf{0.024} \\
\midrule
\multicolumn{5}{l}{\textit{CHD Component Configuration}} \\
CHD-1D Only  & 0.128 & 0.038 & 0.024 & 0.032 \\
CHD-2D Only & 0.135 & 0.041 & 0.027 & 0.035 \\
\rowcolor{blue!10}CHD-1D + CHD-2D & \textbf{0.112} & \textbf{0.030} & \textbf{0.017} & \textbf{0.024} \\
\midrule
\multicolumn{5}{l}{\textit{Codebook Size}} \\
1024 & 0.142 & 0.041 & 0.031 & 0.037 \\
2048 & 0.128 & 0.036 & 0.024 & 0.031 \\
\rowcolor{blue!10}4096  & 0.112 & \textbf{0.030} & \textbf{0.017} & 0.024 \\
8192 & \textbf{0.109} & 0.031 & \textbf{0.017} & \textbf{0.022} \\
\midrule
\multicolumn{5}{l}{\textit{Choice of Distance Metric}} \\
Cosine Distance & 0.135 & 0.038 & 0.025 & 0.031 \\
Wasserstein Distance & 0.148 & 0.041 & 0.029 & 0.035 \\
KL Divergence & 0.124 & 0.033 & 0.021 & 0.028 \\
\rowcolor{blue!10}Hellinger Distance  & \textbf{0.112} & \textbf{0.030} & \textbf{0.017} & \textbf{0.024} \\
\midrule
\multicolumn{5}{l}{\textit{Token Sequence Length}} \\
32 tokens   & 0.157 & 0.042 & 0.043 & 0.050  \\
64 tokens   & 0.140 & 0.037 & 0.028 & 0.034 \\
\rowcolor{blue!10}128 tokens  & \textbf{0.112} & \textbf{0.030} & \textbf{0.017} & \textbf{0.024} \\
\midrule
\multicolumn{5}{l}{\textit{Input Resolution}} \\
$224 \times 224$ & 0.119 & 0.032 & 0.019 & 0.027 \\
\rowcolor{blue!10}$256 \times 256$ & \textbf{0.112} & \textbf{0.030} & \textbf{0.017} & \textbf{0.024} \\
$512 \times 512$ & 0.125 & 0.035 & 0.022 & 0.030 \\
\midrule
\multicolumn{5}{c}{\textit{CMMS / Acc $\uparrow$}} \\
\midrule
\multicolumn{5}{l}{\textit{Input Representation}} \\
Input Images (Pixel-based) & 67.8 & 69.0 & 58.4 & 62.2 \\
\rowcolor{blue!10}Input Tokens  & \textbf{71.5} & \textbf{74.9} & \textbf{61.3} & \textbf{66.7} \\
\midrule
\multicolumn{5}{l}{\textit{Quality Mapping Function $q(p)$}} \\
Linear: $q(p) = 1 - p$ & 68.3 & 71.2 & 57.8 & 63.1 \\
Polynomial: $q(p) = (1-p)^3$ & 69.7 & 72.8 & 59.5 & 64.8 \\
$\exp(-10p)$ & 70.2 & 73.5 & 60.1 & 65.3 \\
\rowcolor{blue!10}$\exp(-20p)$ & \textbf{71.5} & \textbf{74.9} & \textbf{61.3} & \textbf{66.7} \\
$\exp(-30p)$ & 70.9 & 74.1 & 60.7 & 66.0 \\
\midrule
\multicolumn{5}{l}{\textit{Training Strategy}} \\
Token Corruption Only & 69.7 & 72.2 & 57.0 & 61.4 \\
Pixel Augmentation Only & 70.1 & 73.7 & 59.8 & 64.2 \\
\rowcolor{blue!10}Both Combined & \textbf{71.5} & \textbf{74.9} & \textbf{61.3} & \textbf{66.7} \\
\bottomrule
\end{tabular}%
}
\label{tab:ablation_study}
\end{table}

\subsection{Experimental Results}

\textbf{Correlation with human judgments.}
We first assess how well CHD and CMMS track human quality ratings on AGIQA and HPDv3.  
As shown in Table~\ref{tab:performance_comparison4} and Table~\ref{tab:generation_metrics}, CHD achieves Spearman correlations of 0.829 on AGIQA and 0.867 on HPDv3, outperforming distribution-based metrics such as FID, KID, CLIP-FID, DINO-FID, and CMMD, while also attaining the lowest N-MSE.  
CMMS further improves alignment with human scores: it reaches $\rho=0.943$ and N-MSE of 0.050 on AGIQA, and $\rho=0.872$ on HPDv3, consistently surpassing IQA baselines such as MUSIQ, CLIP-IQA, QualiCLIP, and DeQA-Score.

\textbf{Pairwise preference prediction.}
We next evaluate CMMS on binary preference prediction across AGIQA, HPDv2, HPDv3, and VisForm.  
Table~\ref{tab:pref_models_datasets} shows that CMMS achieves the best accuracy on all four benchmarks, with 71.5\% on AGIQA, 74.9\% on HPDv2, 61.3\% on HPDv3, and 66.7\% on VisForm.  
CMMS consistently outperforms recent preference and quality models including QualiCLIP, MDIQA, and DeQA-Score, indicating that token-based representations are effective not only for absolute quality but also for fine-grained human preference modeling.

\textbf{Robustness on VisForm.}
We analyze robustness and generalization on the VisForm benchmark, which spans diverse models and visual domains.  
Figure~\ref{fig:visform} summarizes metric–human correlations across 12 generative models (left) and 21 visual domains (right).  
CHD maintains high correlations in both views, with average Spearman/Kendall of $(0.93, 0.89)$ across models and $(0.87, 0.73)$ across domains, including medically oriented, artistic, and abstract categories.  
In contrast, traditional pixel-based metrics such as FID exhibit pronounced performance drops on non-photorealistic domains (e.g., sketches, collages), suggesting that token histograms capture more domain-agnostic structure.

\begin{figure}[t!]
\begin{center}
\includegraphics[width=0.8\linewidth]{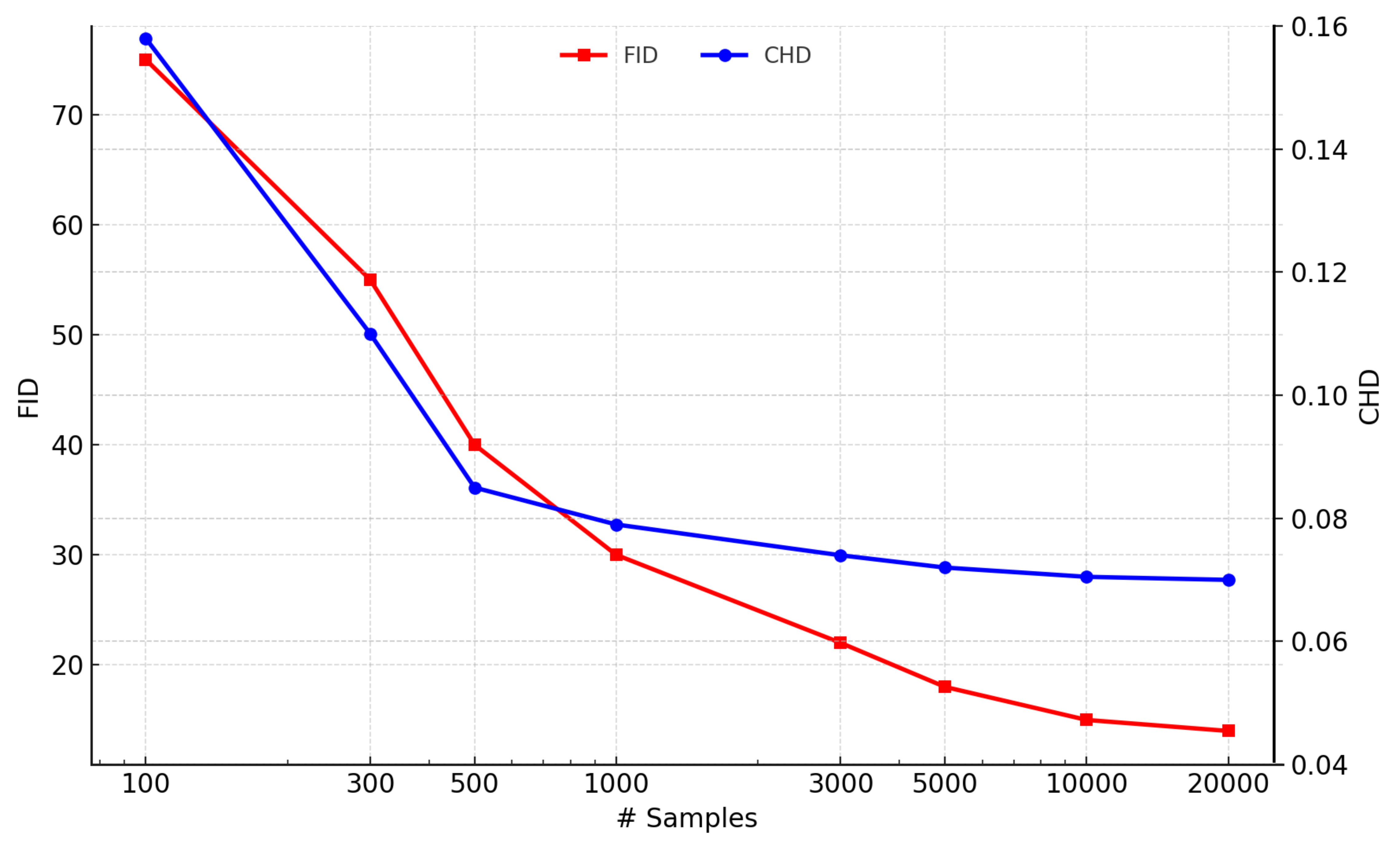}
\end{center}
\caption{Mean CHD and FID values versus sample size. CHD converges with roughly 1{,}000 images, while FID needs over 10{,}000 samples to stabilize.}
\label{fig:exp1}
\end{figure}

\textbf{Sample efficiency.}
Finally, we compare sample efficiency.  
Figure~\ref{fig:exp1} plots the mean CHD and FID values as a function of the number of samples.  
CHD stabilizes with around 1{,}000 images, makes CHD more suitable for evaluating expensive models or limited-sample regimes.

\subsection{Ablation Study}

We ablate key design choices for both CHD and CMMS (Table~\ref{tab:ablation_study}). One-dimensional tokenizers (Instella-T2I, TiTok) significantly outperform 2D tokenizers (VQ-VAE, VQGAN), confirming the advantage of 1D code sequences for distribution matching.  
Combining unigram and 2D co-occurrence statistics (CHD-1D+2D) consistently yields the lowest N-MSE, as the former captures global vocabulary usage and the latter encodes local grammar.  
Performance generally improves with codebook size, with 4{,}096 entries providing a good trade-off and only marginal gain beyond 8{,}192.  
Among distance metrics, Hellinger distance achieves the best overall performance, likely due to its symmetry and bounded range.  
Using 128 tokens at $256\times256$ resolution offers the best balance between detail and efficiency; shorter sequences underfit, while higher resolutions bring limited gains but higher computational cost.

For CMMS, using discrete tokens as input clearly outperforms pixel-based features on all benchmarks, improving preference accuracy by 3.7–5.9 points.  
The exponential mapping $q(p)=\exp(-20p)$ provides the best calibration between corruption level and target quality, outperforming linear and polynomial alternatives.  
Finally, combining token corruption with pixel-space augmentations yields the strongest results: either source alone leads to noticeable drops in accuracy, confirming that the two degradation families provide complementary supervision for learning robust perceptual scores.

\vspace{-6pt}

\section{Conclusion}

We present a discrete token-based framework for generative model evaluation by shifting from continuous features to codebook statistics. Based on this paradigm, we develop two complementary metrics: CHD for distribution matching and CMMS for reference-free quality assessment. Both achieve state-of-the-art correlation with human judgment across multiple benchmarks, including VisForm. Our framework is scalable, interpretable, and robust to domain shifts, offering a unified solution for perceptually aligned quality assessment. Future work will explore higher-order token statistics for improved spatial modeling, as well as extensions to video and 3D generation.

\section*{Acknowledgements}
This work was completed at the Pattern Recognition Center, WeChat AI, Tencent. The authors would like to express their sincere gratitude to Professor Anil K. Jain for his valuable guidance, insightful discussions, and continuous encouragement, which greatly benefited this work.
{
    \small
    \bibliographystyle{ieeenat_fullname}
    \bibliography{main}
}


\end{document}